# Optimization of Precipitate Segmentation Through Linear Genetic Programming of Image Processing

Kyle Anthony Williams,[a]* Andrew H. Seltzman[b]

[a] Massachusetts Institute of Technology, Electrical Engineering and Computer Science, Cambridge, MA; https://orcid.org/0009-0001-3193-328X

[b] Massachusetts Institute of Technology, Plasma Science and Fusion Center, Cambridge, MA; https://orcid.org/0000-0002-7725-2981

*E-mail: kawcco@mit.edu

# Optimization of Precipitate Segmentation Through Linear Genetic Programming of Image Processing

Neutron flux in fusion reactors causes voids and swelling in materials—and creates long-lived nuclear waste in certain material isotopes susceptible to neutron activation. Current additive manufactured niobium-based copper alloys such as GRCop-84/42 have high yield-strength due to precipitates that prevent swelling by acting as sinks for helium generation and displaced atoms. Neutron activation of niobium produces long-lived isotopes; new reduced-activation copper alloys with similar yield-strength are needed. The yield-strength of a material can be estimated through analysis of precipitate size distribution from electron micrographs of focused-ion-beam (FIB) milled cross-sections. Current analysis relies on hand annotation due to varying contrast, noise, and image artifacts present in micrographs, slowing iteration speed in alloy development. We present a filtering and segmentation algorithm for detecting precipitates in FIB cross-section micrographs, optimized using linear genetic programming (LGP), which accounts for the various artifacts. To this end, the optimization environment uses a domain-specific language for image processing to iterate on solutions. Programs in this language are a list of image-filtering blocks with tunable parameters that sequentially process an input image, allowing for reliable generation and mutation by a genetic algorithm. Our environment produces optimized human-interpretable MATLAB code representing an image filtering pipeline. Under ideal conditions—a population size of 60 and a maximum program length of 5 blocks—our system was able to find a near-human accuracy solution with an average evaluation error of 1.8% when comparing segmentations pixel-by-pixel to a human baseline using an XOR error evaluation. Our automation work enabled faster iteration cycles and furthered exploration of the material composition and processing space: our optimized pipeline algorithm processes a 3.6 megapixel image in ~2 seconds on average. This ultimately enables convergence on strong, low-activation, precipitation hardened copper alloys for additive manufactured fusion reactor parts.

Keywords: image processing; genetic programming; machine learning; fusion materials; metallurgy

## I. INTRODUCTION

Additive manufacturing (AM) is a key enabling technology in the rapid development of radio-frequency structures for fusion reactors [1] [2]. The harsh fusion reactor environment exposes AMed parts to high temperatures and disruption loads, which can cause parts to soften and warp. Cu-Cr-Nb alloys are commonly used in AM of reactor parts because of their resistance to warping. However, niobium is susceptible to neutron activation into long-lived isotopes, which results in nuclear waste generation after prolonged exposure [3]. New reduced-activation alloys with high tensile strength and conductivity are needed; we focus on the former attribute in this communication.

By the Orowan strengthening law, average precipitate radii and overall volume influence the tensile strength in copper alloys: finer distributions of precipitates are correlated to higher tensile strength [4], [5]. Precipitate size distributions can be gathered from electron micrographs of alloy cross-sections [6], [7]. Analysis of electron microscopy (EM) is challenging due to artifacts introduced during image acquisition [8]. Artifacts present in our dataset included non-uniform illumination, blurring, and noise. Further complications arise from the visual properties of our samples that feature non-uniform background contrast between varying copper grain orientations and $Cr_2Nb$ precipitates, in addition to sharp transitions between grain boundaries, and streaking lines introduced by the focused ion beam milling process used to create cross-sections. More rigorous qualitative and quantitative models of degradation can be found in [8] and [9], respectively. Our group currently processes material micrographs via computer-assisted hand annotation which, despite high accuracy, can take upwards of an hour for a single image: automation of this is needed.

**[Figure 1]**

The complex nature of micrograph denoising and deconvolution makes it challenging for a human to devise a filter-segmentation algorithm for micrographs of material cross-sections that is both accurate and general. Therefore, it is common to hand off the design of an algorithm to the computer through machine learning. Much has been written on using traditional deep learning architectures to fully automate solving materials-specific segmentation problems [10], [11], and new architectures supported by transformers look to minimize the need for training data. [12], [13], [14], [15] However, much work in this area fails to consider the growing importance of interpretability of machine learning models. [16] In our case: why and how did a model choose to segment a part of the image? If a model makes a mistake, how can I use my understanding to right its mistake? It's especially important that this reasoning can be put in terms relevant to the domain of materials scientists, the ultimate end-users of these models. It is well-known that large deep learning architectures like those cited above are very challenging to interpret. In this paper, we hope to develop successful segmentation solutions that are relatively easy to interpret and edit.

In this paper, we focus on machine learning to optimize image-processing for precipitate size analysis in prospective materials. We introduce linear genetic programming (LGP) as an automated way of discovering effective image processing *pipelines*, algorithms built from applying smaller algorithms in sequence, for processing our dataset. LGP allows us to focus on just defining the solution space to the problem of filtering out the various artifacts, an arguably much simpler task thanks to existing literature. An environment for generating and optimizing pipelines was created in MATLAB. We found the environment was able to produce pipelines with near-human accuracy. Furthermore, the environment demonstrated emergent behavior as population size and pipeline length was increased, including overcoming failure modes and recreating aspects of crucial nonlinear operations.

# II. BACKGROUND

## II.A. Linear Genetic Programming

Evolutionary algorithms (EAs) are a class of optimization techniques centered around aspects of natural selection and evolution. Possible solutions to a problem are randomly generated, evaluated, ranked, and evolved over the course of multiple iterations, or *generations*, until a satisfactory solution is found. We refer to an implementation of EAs as an *environment*: the problem as defined by the computer, the solution space, and how to navigate it through evaluation and evolution algorithms. The key differentiator between different evolutionary algorithms is how solutions are represented and evolved. Genetic programming (GP) represents solutions as computer programs constructed from a set of predefined *building blocks*. The process of GP is as follows, as adapted from Algorithm 1.1 of [17]:

1. Randomly generate a *population*, or set of programs; they will act as the first generation.
2. Rank the generation according to *fitness*, or how well each program emulates the human-based solution evaluated based on a given metric.
3. The highest fitness programs create *children* programs, a process called *crossover,* where the blocks in the child program pipelines are combined from the parent programs.
4. Randomly *mutate* certain attributes of those children, changing the parameters of each filter block.

5. Repeat from (3) until a stopping criteria has been met, such as a maximum number of generations being reached, or a certain measure of fitness.
6. Present the program with the highest fitness.

**[Figure 2]**

The key differentiator between different forms of GP is how programs and their building blocks are represented and evolved. Our method of choice, *linear genetic programming*, represents programs as a sequence of instructions, in our case image processing algorithms. Traditionally, LGP is achieved by having a computer evolve the programs written for a register-based virtual machine or imperative programming language. To ensure simplicity of implementation and that correct programs are always generated, our method differs from the literature and represents our programs as one-dimensional graphs. Each instruction in our set has one input and one output, and can be tuned by a series of constant parameters. This is in contrast to the literature [18], which features instructions of variable arity. A program, then, is simply a sequence of these instructions:  input data implicitly enters the first block, and its output is the input of the second. This is a paradigm known as *tacit programming*, and is regularly seen in software engineering to represent pipelines. This is similar in premise to the other two popular forms of program representation in GP— *tree-based* GP and *Cartesian* GP—which also represent programs as directed acyclic graphs, but with significantly more restrictions. Tree-based GP represents programs as nested expressions, while Cartesian GP represents programs as a graph where nodes have multiple inputs and outputs, usually visualized on a 2D plane. Further information on evolutionary algorithms and (especially linear) genetic programming is found in [17].

### II.B. Micrograph Processing

The kind of computer vision problem that we solve in this communication is known as *image restoration* [19]. In this communication, the word “image” is a shorthand for a 2D grayscale image of arbitrary finite size. An image $I$ of size $m \times n$ can be represented as a matrix of natural numbers ranging from 0 to 255 inclusive. We solve for some *real image* $I$ in the equation $I' = T(I)$, where $T$ is some process that modifies *I*, and *I’*, the modified image filled with artifacts, is our input. To do this, we are trying to find $T^{-1}$, the inverse of the process $T$ such that $T^{-1}(I') = I$. The concept of the real image is abstract: because the artifacts in our image were not synthetically added to a known image, the exact reconstruction target is unknown. Instead, we focus on identifying specific issues that are common among our dataset, and introduce objectives and instructions that in theory should address those issues. In this sense, the real image is a reduction of the artifacts present in $I'$ such that a precipitate size distribution can be determined. Image restoration in our work and in the literature more broadly sits at the intersection of many subdisciplines of computer vision, most notably denoising (removing noise from an image) and deconvolving (inverting a class of operations that includes effects like blurring).

To guide our restoration work, we have adopted the degradation models posed by Joris Roels [8], [9], while making necessary extensions to fit our data. Most of the degradations originate from the microscopy process, primarily from the nature of sample interaction itself:

1. *non-uniform illumination* of the sample—some regions are darker than others due to shielding of secondary electron emission near the bottom of FIB cross-sections—can be resolved with flat-field correction, which selectively illuminates darker parts of the image
2. *blurring* due to the varying focal length along the y-axis of the scan can be resolved with sharpening and deconvolution tools which re-emphasize edges, and
3. *noise*—usually a mix of Gaussian and Poisson—can be resolved with filters and other denoising algorithms, which average or “smooth” out the noise with surrounding data.

Extra problems arise from sample preparation:

4. Because our samples are alloys, their cross-sections feature *non-uniform background contrast* across different grain orientations with sharp transitions marking the boundaries between grains; this can be resolved with histogram equalization or illumination algorithms.
5. *Streaking lines*, referred to by some in the literature as *stripe noise*, are introduced by the focused ion beam milling process used to create cross-sections; these can be resolved with blurring and filters.

**II.C Genetic Programing Use in Image Processing**

Here we discuss micrograph restoration techniques that have influenced our work. In [9], an overview and comparison of denoising and deblurring techniques are presented, including multiple machine learning solutions (albeit before the transformer boom), in the context of electron microscopy. This work was useful for identifying various well-understood filtering techniques in the EM literature; we then implemented these techniques in our optimization environment as building blocks such that they could then be composed into larger solutions. In [8] and [9], multiple deconvolution techniques to resolve specific artifacting problems were described; these were implemented in our environment. The literature on image restoration has historically focused on applying GP to image denoising and little on deconvolution [20]. The closest parallel we've found to our work is [21], which applies Cartesian GP to various deconvolution-segmentation tasks in biomedical imaging. The technique they showcase is designed for information-dense images, with data as large as high-resolution, full-color, 3D images of cells. They evaluate their programs by translating their representations to just-in-time generated Python code backed by the OpenCV image processing library, an approach very similar to ours. Another parallel to our work, which focuses exclusively on denoising, is [22], which uses tree-based GP to compose well-known filtering operations along with basis arithmetic to compose local adaptive filters for images containing Gaussian and salt-and-pepper noise.

## III. METHODS

### III.A. Environment Methodology

LGP is traditionally achieved by having a computer evolve the programs written for a register-based virtual machine or imperative programming language [1] [2]. To ensure simplicity of implementation and that correct programs are always generated, we represent our programs as one-dimensional graphs. Each block, or instruction, has one input and one output, and can be tuned by a series of constant parameters. This is in contrast to the literature, which features instructions of multiple variable inputs and control flow operators [1] [2]. If programs using the latter paradigm are generated naïvely, this can lead to programs that don't typecheck or never terminate, complicating the ranking process. A program, then, is simply a sequence of these instructions: input data–in our case, a 2D grayscale image–implicitly enters the first block, and its output is the input of the second. This method is known as tacit programming, which is regularly used to represent pipelines, most notably the Unix operating system and its successors [23].

Our system features a library of 14 blocks which wrap around implementations of image processing routines from MATLAB's Image Processing Toolbox[1]. Each features constant parameter inputs. Blocks were chosen for their general use in restoration or their specific application in micrograph restoration according to [8] and [9]. For the functionally inclined, blocks are implemented as higher order functions of the type $(P_1, ..., P_m) \rightarrow I \rightarrow I$, where $P_i$ is the type of the $i$-th parameter (possible types include odd integers, natural numbers, floats, and enumerations), and $I$ is the set of all grayscale 2D images. Once parameters are passed, blocks can then be composed to form pipelines of the type $I \to \ldots \to I$.

[1] For a list of all blocks, see the Supplementary Materials.

New solutions are explored during the optimization process through crossover and mutation. Crossovers affect the order and presence of filter blocks in a child solution. We engage in uniform crossover, which selects blocks at random from two parent solutions. To determine the pairs of parents, we partition the elite into two halves and pair solutions from half with solutions from the other. Mutations act on a block level, affecting only the values of block parameters. Parameters are mutated by sampling normal distribution and discrete uniform distributions as appropriate for the data type of the parameter.

Our environment's fitness function encompasses five objectives that are simultaneously minimized using Pareto optimization, which aims to identify solutions where an improvement in one objective requires a degradation in another metric. A set of solutions which both feature this characteristic and minimize the objective functions as much as possible is called a *Pareto front*. For more information, see [24]. The first four objectives are surrogates related to effective segmentation of the precipitates from the image. For any individual micrograph in the training set, a solution is graded on:

A. its "area error," the difference in area between the LGP solution's segmentation and the human segmentation;
B. "precipitate count error," the difference in the number of precipitates the computer has segmented compared to human segmentation;

C. "precipitate size distribution distance," the norm of the difference between the normalized histograms of precipitates segmented according to their size of the solution and the human segmentation
D. "XOR error," the percentage of pixels that are non-identical between the solution and the human segmentation.

Finally, we also have an objective related to performance, how long it took in seconds for a solution to process an image. For the entire training set, a collective error is determined for each objective using the mean-squared error, and an average "time taken per image squared" is calculated: these are then the "scores" for each objective. These scores, put together in a vector, comprise a solution's *fitness score*.

We list relevant parameters of our environment below:

- MATLAB's multi-objective genetic algorithm [25] is based on NSGA-II [26].
- We use tournament selection with a size of 4 [17].
- Raw fitness score is scaled by $r^{-1/2}$, where $r$ is a solution's "rank" in a generation's population in order to make sure unsuccessful algorithms are more nearly equal in score.
- The $n + 1$-th generation is composed of 80% of crossover children and 5% percent of the most elite solutions of generation $n$.
- We run the optimization for at most 500 generations and give the process a maximum runtime of 20 hours.

**[Figure 3]**

**III.B. Experimental Methodology**

Two independent variables were changed prior to running the optimization process; the generated solutions on the Pareto front were then evaluated with our fitness function against the micrographs in the evaluation set. We use a 75% training, 25% evaluation split with 21 total examples. Images in our dataset are 8-bit grayscale at a resolution of 2048 by 1767.[2] We investigate two independent variables: population sizes of 20, 40, and 60 solutions; and maximum solution lengths of 4, 5, and 6 blocks. Each population/solution length combination is run three times each. Since each trial outputs a Pareto front of seven solutions each, we share the min, max, mean, and median of the statistical performance of the solutions of each trial. In our environment, we only optimize the filtering half of the filter-segmentation process; the segmentation procedure is fixed code written by the second author. All results were found using MATLAB R2024a (version 24.1.0.2537033) [27] using the Image Processing (version 24.1), Statistics and Machine Learning (version 24.1), Parallel Computing (version 24.1), and Global Optimization Toolboxes (Version 24.1) with code commit `10bd5051` on Linux 4.18.0-348.el8.0.2.x86_64 with double-precision floating point arithmetic on 10 vCPUs with 40 GB of RAM. All code, datasets, raw results data, and a data dictionary can be found in the Supplementary Materials.

**IV. RESULTS**

**[Table 1]**

[2] It is important to note that the area of a precipitate in pixels can vary wildly based on a micrograph's magnification and the precipitate's makeup. We hope for segmentation to be robust in a variety of situations.

Throughout this section, we adopt a syntax for identifying specific trials: the identifier "*x*-*y*-*z*" points at the *z*-th trial with a population size of *x* and a maximum program length of *y* blocks. For example, the identifier "20-4-1" points at the first trial with a population size of 20 and maximum program length of 4 blocks.

Of the 27 trials, solutions from the 60-5-*z* trials minimized average XOR error across our evaluation set, with trial 3 featuring a solution with an average error of 1.79%. Solutions generally tried using as few blocks as possible. Populations tended to randomly oscillate around an average generation distance, the average of the loss for each solution in a generation, of 0, but with smaller amplitudes over time. For most trials, evaluating a single generation would take less than 30 minutes; many trials would initially feature large generation durations that shrank and eventually plateaued as solutions matured. For popular blocks with parameters, programs would converge onto small intervals within a few dozen generations.

Each cell of **[Table 1]** represents the performance of the best solution in each trial's Pareto front. We take an average of averages to make more obvious associations between population size, program length, and performance. Larger population size is associated with smaller error. A maximum program length of 5 blocks minimizes error. A population of 60 and program length of 5 was the most successful; the average trial error is 1.8%, with trial 3 featuring a solution with an average error of 1.79%.

**[Figure 4]**

**[Figure 5]**

**[Figure 6]**

When using the best performing pipeline from 60-5-3 to create a precipitate size distribution, its work is comparable to that of the human baseline. The pipeline's distributions generally have wider interquartile ranges and less outliers; its median is usually less than that of the human, likely due to segmentation errors producing small blobs which skew the median downward.

**[Figure 7]**

A description of the pipeline can be found below:

1. `imadjust()`
2. `adapthisteq(15, 21)`
3. `imlocalbrighten(0.63579, false)`
4. `imsharpen(37.3409, 1.363, 0.28738)`
5. `medfilt2(5, 7, zeros)`

MATLAB code representing the above pipeline:

```
pipeline(Blocks.imadjust(), Blocks.adapthisteq(15, 21),
Blocks.imlocalbrighten(0.63579, false),
Blocks.imsharpen(37.3409, 1.363, 0.28738), Blocks.medfilt2(5,
7, 'zeros'))
```

## V. DISCUSSION

### V.A. Interpretation of Results

**[Figure 8]**

**[Figure 8]** conveys the general timeline of block popularity in our environment. The heavy use of the identity block implies that shorter pipelines were preferred, as previously noted. Computationally expensive blocks like the Gaussian filter (`imgaussfilt_square`) and the local Laplacian (`locallapfilt`) "die out" quickly. No one block finds stability, but blocks like `histeq` (histogram equalization) fall in popularity as the environment discovers more powerful alternatives like `adapthisteq` (adaptive histogram equalization). Most actively used blocks vary in their frequency stochastically because of uniform crossover.

**[Figure 9]**

Our environment demonstrates convergent behaviour as pipeline length and population size is increased. On smaller population and pipeline lengths, segmentations of precipitates overlapping with streaks become inaccurate, leaving large blotches and streaks where streaks are located. We find that this behaviour disappears at a pipeline length of 5 and population of 60. Looking back at [table error], larger population sizes look to have a clear effect on minimizing error. We recommend that the reader make their population size as large as their compute allows.

Our environment emulated well-understood effects without making direct use of them. Elements of the 60-5-3 pipeline are similar to flat-field correction [28]. In **[Figure 9]**, the back-to-back histogram equalizations (`imadjust` and `adapthisteq`) amplifies the noise of the image and raises the contrast of the precipitates. Afterwards, the low-level image brightening (`imlocalbrighten`) eliminates background contrast of the alloy. Finally, the unsharp mask (`imsharpen`) enhances precipitate outlines before the median filter (`medfilt2`) cleans up salt-and-pepper pixel noise and leaves the precipitates as the darkest objects on the surface of the cross-section. In all, the dark frame is subtracted from the raw image, and then averaged out, a core part of flat-field correction.

The pipeline shows how our block naturally addresses Roels et al.'s model of micrograph artifacts [8]. Non-uniform illumination is handled by (an emulation of) flat-field correction, blur is handled by the unsharp masking, and noise is handled by the median filter. It also handles artifacts specific to our model: background contrast is handled with histogram equalization and local brightening, and streaking is handled by the median filter. As with any machine learning project, it is of note that our fitness functions make no mention of the specific artifacts. The environment, instead, naturally composed effective pipelines using tools built for the task simply guided by segmentation accuracy compared to a human.

**[Figure 10]**

Focusing in on `adapthisteq`, we can see how the environment optimized the neighborhood height but experimented with neighborhood height throughout the trial, as seen in **[Figure 10]**. Convergence on some block parameters and wide spreads on others is a common theme in the environment.

Our solutions feature some common failure modes that are not overcome during the optimization process. The most notable are clumping close precipitates together in the segmentation, and missing small precipitates all together. Streaking, while subdued as population size and pipeline length is increased, is still not fully resolved; our segmentation algorithm can still confuse streaks for precipitates, which can make precipitate distributions more noisy.

**[Figure 11]**

**[Figure 12]**

Despite the above failure modes, we still argue that the outputs of our process can reasonably be described as near-human. Quantitatively, as seen in **[Figure 9]**, the outputs of the 60-5-3 pipeline do a good job of mimicking the qualities of a human segmentation at a glance, which has been found to be an important part of segmentation evaluation. [29] Furthermore, it could be argued that, on some views, our approach results in more accurate segmentation than other humans when compared to a human baseline: [30] found that that individual human annotators, regardless of prior experience, on average had a Jaccard distance of about 28% (lower is better), a metric closely related to our XOR error.

### V.II. Design Outcomes

In contrast to contemporary machine learning techniques, the solutions produced by our environment are much lighter in multiple regards. Because our solutions are an abstract representation of function calls, as opposed to a neural network's large collection of weights, we are able to directly translate solutions into MATLAB code that can be executed with a minimal runtime. Once in this form, a solution can then be further interpreted and hand-optimized by a human to fit specific use cases: a form of interpretable machine learning [16]. This is possible because we create solutions from complex, but well-understood filters from the literature, instead of composing entirely novel filters from more foundational operators. Papers like [31] and [21] approach our technique by including higher-order operations in their libraries. Yan *et al.* notably incorporate multiple parameterizations of the bilateral filter. Cortacero et al. incorporate a series of blurring, sharpening, and specialized morphological operations. However, both still incorporate operators like matrix addition and bitwise masking; we avoid these operations to make interpretation as simple as possible.

Our environment also makes reasoning about the design significantly easier than contemporary solutions. The library of blocks is easily expanded and edited as previous solutions become less effective with additional data. By generating programs in a declarative domain-specific language, as opposed to an imperative language; we avoid entire classes of correctness problems like non-finite runtime programs [17] due to implementing the design principles of functional programming: pure functions are a natural way to represent image processing operations, and their composition is trivially reasoned about. We've achieved this without sacrificing performance: our 60-5-3 pipeline processes a micrograph (~3.6 megapixels) in ~2 seconds on a standard workstation, orders of magnitude faster than a human hand-annotating a single image.

## V.III. Future Work

The next step for our work is validating it against new materials unseen in our training set. For the sake of time, we plan to handle this in a future communication.

While we were able to mitigate artifact-induced errors present in our data set, LGP filters were not optimal for streaks (similar effects are referred to as *stripe noise* in the literature) due to the absence of a block directly addressing streak artifacts. General procedures for removing stripe noise are well-understood [32], but require fine-tuning to work for a specific dataset, something that our LGP technique is prepared to address.

[21]. An open question is how much control the environment should have over the procedure design. There is no additional image processing before filtering and in-between filtering and segmentation; the environment has full control over the process. With segmentation, giving the environment full rein could be computationally

wasteful: leaving it to discover that, for example, filling holes in segmentations is a good idea, when we already know that, will undoubtedly make the optimization process longer. A good approach may be to have the optimization start from a few mutations of an existing pipeline and optimize it from there.

We would like to experiment with different mutation and crossover algorithms and see which ones are better, if any, for our use case than uniform crossover and our “parameter mutation” algorithms. A possible limitation of how we implement mutation, for example, is that if a block is not part of any member of the initial population or “dies out” in a later generation, it will never show up in a solution, which may shrink our solution space too early. Ultimately, the algorithms currently used stuck because they’re simple and they produced satisfactory results, not necessarily because they’re most appropriate for our use case

Finally, how to pick a solution from the Pareto front, while well discussed in the literature [33], [34], is not something we’ve looked into at the time of writing. While our objectives are mostly complementary; picking a solution by picking a single fitness metric, and picking the solution that does best on the validation set by that metric, remains to be examined.

## VI. CONCLUSION

In this paper, we present a machine learning environment inspired by linear genetic programming that was able to reach near-human accuracy on precipitate segmentation in cross-sections of additive manufacturing powders. Using a declarative, functional domain-specific language for creating image processing pipelines from well-understood algorithms, emergent properties, including the ability to replicate algorithms commonly used in the literature and overcome specific failure modes, were found as population size and pipeline length were increased. Solutions are easily interpreted and integrated into codebases by consumers. Effective filter-segmentation programs produced will save tens of man hours per alloy analysed, enabling rapid discovery of materials appropriate for use in the additive manufacturing of fusion reactor parts like RF waveguides. This work shows that evolutionary methods can be easily integrated into the data processing workflows of materials science for a wide variety of applications and more easily reasoned with than neural networks.

## VII. ACKNOWLEDGEMENTS

We would like to thank the following people for assistance with this communication: Alexis Devitre, Myles Stapelberg, Brianna Ryan, Alejandro Paz and Phoebe Ayers (MIT Libraries), Tim Mattson, and Rachel Shulman.

We thank the MIT PSFC's Fusion and Fission Undergraduate Scholars program and the MIT Undergraduate Research Opportunities Program for funding and supporting this work. Work supported by US DOE under DE-SC0014264. This work made use of the MRSEC Shared Experimental Facilities at MIT, supported by the National Science Foundation under award number DMR-14-19807.

## VIII. DISCLOSURE STATEMENT

The authors declare that they have no known competing financial interests or personal relationships that could have appeared to influence the work reported in this paper.

## IX. DATA AVAILABILITY

MIT-licensed source code and data can be found at https://github.com/kawcco/find-the-precipitates.

## REFERENCES[3]

[1] Seltzman, A. H., and S. J. Wukitch. 2021. Resolution and Geometric Limitations in Laser Powder Bed Fusion Additively Manufactured GRCop-84 Structures for a Lower Hybrid Current Drive Launcher. *Fusion Engineering and Design* 173 (December):112847. doi:10.1016/j.fusengdes.2021.112847.

[2] Seltzman, A.H., S. Shiraiwa, G.M. Wallace, and S.J. Wukitch. 2019. A High Field Side Multijunction Launcher with Aperture Impedance Matching for Lower Hybrid Current Drive in DIII-D Advanced Tokamak Plasmas. *Nuclear Fusion* 59 (9). IOP Publishing:096003. doi:10.1088/1741-4326/ab22c8.

[3] Wallace, Gregory M., Elena Botica Artalejo, Michael P. Short, and Kevin B. Woller. 2022. Ultra-Rapid, Physics-Based Development Pathway for Reactor-Relevant RF Antenna Materials. *IEEE Transactions on Plasma Science* 50 (11):4506–4509. doi:10.1109/TPS.2022.3171497.

[4] Seltzman, A. H., and S. J. Wukitch. 2021. Fracture Characteristics and Heat Treatment of Laser Powder Bed Fusion Additively Manufactured GRCop-84 Copper. *Materials Science and Engineering: A* 827 (October):141690. doi:10.1016/j.msea.2021.141690.

[5] Gladman, T. 1999. Precipitation Hardening in Metals. *Materials Science and Technology* 15 (1). SAGE Publications:30–36. doi:10.1179/026708399773002782.

[6] Seltzman, A. H., and S. J. Wukitch. 2021. Precipitate Size in GRCop-84 Gas Atomized Powder and Laser Powder Bed Fusion Additively Manufactured

[3] For reviewers: unfortunately, our citation software (Zotero) does not allow us to have separate citation numbers for the same document, which is required to satisfy this journal's style needs. We hope to resolve this manually at the end of the review process. Thank you in advance for understanding.

Material. *Fusion Science and Technology* 77 (7–8). American Nuclear Society:641–646. doi:10.1080/15361055.2021.1913030.

[7] Seltzman, A. H., and S. J. Wukitch. 2023. Precipitate Size in GRCop-42 and GRCop-84 Cu-Cr-Nb Alloy Gas Atomized Powder and L-PBF Additive Manufactured Material. *Fusion Science and Technology* 79 (5). Taylor & Francis:503–516. doi:10.1080/15361055.2022.2147765.

[8] Roels, Joris, Jan Aelterman, Jonas De Vylder, Saskia Lippens, Hiêp Q. Luong, Christopher J. Guérin, and Wilfried Philips. 2016. Image Degradation in Microscopic Images: Avoidance, Artifacts, and Solutions. In *Focus on Bio-Image Informatics*, ed. Winnok H. De Vos, Sebastian Munck, and Jean-Pierre Timmermans, 41–67. Cham: Springer International Publishing. doi:10.1007/978-3-319-28549-8_2.

[9] Roels, J., J. Aelterman, H. Q. Luong, S. Lippens, A. Pizurica, Y. Saeys, and W. Philips. 2018. An Overview of State-of-the-Art Image Restoration in Electron Microscopy. *JOURNAL OF MICROSCOPY* 271 (3). Hoboken: Wiley:239–254. doi:10.1111/jmi.12716.

[10] Oostrom, Marjolein, Alex Hagen, Nicole LaHaye, and Karl Pazdernik. 2025. Bayesian SegNet for Semantic Segmentation with Improved Interpretation of Microstructural Evolution during Irradiation of Materials. *Computational Materials Science* 257 (July):113943. doi:10.1016/j.commatsci.2025.113943.

[11] Zhou, Peng, Xinyi Zhang, Xuejing Shen, Hui Shi, Jinglin He, Yifei Zhu, Fan Jiang, and Fangzhou Yi. 2024. Multi-Phase Material Microscopic Image Segmentation for Microstructure Analysis of Superalloys via Modified U-Net and Rectify Strategies. *Computational Materials Science* 242 (June):113063. doi:10.1016/j.commatsci.2024.113063.

[12] Li, Changtai, Xu Han, Chao Yao, Yu Guo, Zixin Li, Lei Jiang, Wei Liu, Haiyou Huang, Huadong Fu, and Xiaojuan Ban. 2025. A Novel Training-Free Approach to Efficiently Extracting Material Microstructures via Visual Large Model. *Acta Materialia* 290 (May):120962. doi:10.1016/j.actamat.2025.120962.

[13] Docherty, Ronan, Antonis Vamvakeros, and Samuel J. Cooper. 2025. Upsampling DINOv2 Features for Unsupervised Vision Tasks and Weakly Supervised Materials Segmentation. arXiv. doi:10.48550/arXiv.2410.19836.

[14] Zou, Xueyan, Jianwei Yang, Hao Zhang, Feng Li, Linjie Li, Jianfeng Wang, Lijuan Wang, Jianfeng Gao, and Yong Jae Lee. 2023. Segment Everything Everywhere All at Once. *Advances in Neural Information Processing Systems* 36 (December):19769–19782. https://proceedings.neurips.cc/paper_files/paper/2023/hash/3ef61f7e4afacf9a2c5b71c726172b86-Abstract-Conference.html.

[15] Kirillov, Alexander, Eric Mintun, Nikhila Ravi, Hanzi Mao, Chloe Rolland, Laura Gustafson, Tete Xiao, Spencer Whitehead, Alexander C. Berg, Wan-Yen Lo, et al. 2023. Segment Anything. In *2023 IEEE/CVF International Conference on Computer Vision (ICCV)*, 3992–4003. doi:10.1109/ICCV51070.2023.00371.

[16] Murdoch, W. James, Chandan Singh, Karl Kumbier, Reza Abbasi-Asl, and Bin Yu. 2019. Definitions, Methods, and Applications in Interpretable Machine Learning. *Proceedings of the National Academy of Sciences* 116 (44).

Proceedings of the National Academy of Sciences:22071–22080. doi:10.1073/pnas.1900654116.

[17] Markus F. Brameier and Wolfgang Banzhaf. 2007. *Linear Genetic Programming*. Genetic and Evolutionary Computation. Boston, MA: Springer US. doi:10.1007/978-0-387-31030-5.

[18] Oltean, Mihai, and Crina Groşan. 2024. A Comparison of Several Linear Genetic Programming Techniques. *Complex Systems* 14 (4):285–313. doi:10.25088/ComplexSystems.14.4.285.

[19] Banham, M.R., and A.K. Katsaggelos. 1997. Digital Image Restoration. *IEEE Signal Processing Magazine* 14 (2):24–41. doi:10.1109/79.581363.

[20] Khan, Asifullah, Aqsa Saeed Qureshi, Noorul Wahab, Mutawarra Hussain, and Muhammad Yousaf Hamza. 2021. A Recent Survey on the Applications of Genetic Programming in Image Processing. *Computational Intelligence* 37 (4):1745–1778. doi:10.1111/coin.12459.

[21] Cortacero, Kévin, Brienne McKenzie, Sabina Müller, Roxana Khazen, Fanny Lafouresse, Gaëlle Corsaut, Nathalie Van Acker, François-Xavier Frenois, Laurence Lamant, Nicolas Meyer, et al. 2023. Evolutionary Design of Explainable Algorithms for Biomedical Image Segmentation. *Nature Communications* 14 (1). Nature Publishing Group:7112. doi:10.1038/s41467-023-42664-x.

[22] Yan, Ruomei, Ling Shao, Li Liu, and Yan Liu. 2014. Natural Image Denoising Using Evolved Local Adaptive Filters. *Signal Processing*, Image Restoration and Enhancement: Recent Advances and Applications, 103 (October):36–44. doi:10.1016/j.sigpro.2013.11.019.

[23] Ritchie, D. M. 1984. The UNIX System: The Evolution of the UNIX Time-Sharing System. *AT&T Bell Laboratories Technical Journal* 63 (8):1577–1593. doi:10.1002/j.1538-7305.1984.tb00054.x.

[24] Emmerich, Michael T. M., and André H. Deutz. 2018. A Tutorial on Multiobjective Optimization: Fundamentals and Evolutionary Methods. *Natural Computing* 17 (3):585–609. doi:10.1007/s11047-018-9685-y.

[25] The MathWorks Inc. 2024. Gamultiobj Algorithm. Accessed November 22. https://www.mathworks.com/help/gads/gamultiobj-algorithm.html.

[26] Deb, Kalyanmoy. 2001. *Multi-Objective Optimization Using Evolutionary Algorithms*. Chichester, England: John Wiley & Sons, Ltd. https://www.wiley.com/en-us/Multi-Objective+Optimization+using+Evolutionary+Algorithms-p-9780471873396.

[27] The MathWorks Inc. 2024. MATLAB Version: 24.1.0 (R2024a). Natick, Massachusetts, United States: The MathWorks Inc. https://www.mathworks.com.

[28] Seibert, James Anthony, John M. Boone, and Karen K. Lindfors. 1998. Flat-Field Correction Technique for Digital Detectors. In *Medical Imaging 1998: Physics of Medical Imaging*, 3336:348–354. SPIE. doi:10.1117/12.317034.

[29] Haxhimusa, Yll R. 2022. A Study on Human Image Segmentation for Evaluation of Segmentation Methods. *IFAC-PapersOnLine*, 21st IFAC Conference on Technology, Culture and International Stability TECIS 2022, 55

(39):270–275. doi:10.1016/j.ifacol.2022.12.033.
[30] Blushtein-Livnon, Roni, Tal Svoray, and Michael Dorman. 2025. Performance of Human Annotators in Object Detection and Segmentation of Remotely Sensed Data. *IEEE Transactions on Geoscience and Remote Sensing* 63:1–16. doi:10.1109/TGRS.2025.3555235.
[31] Yang, Chaoming, Mingfei Zhang, and Liang Qi. 2020. Grain Boundary Structure Search by Using an Evolutionary Algorithm with Effective Mutation Methods. *Computational Materials Science* 184 (November):109812. doi:10.1016/j.commatsci.2020.109812.
[32] Ahmed Hamadouche, Sid, Ayoub Boutemedjet, and Azzedine Bouaraba. 2024. Efficient and Robust Techniques for Infrared Imaging System Correction. *The Imaging Science Journal* 0 (0). Taylor & Francis:1–20. Accessed December 5. doi:10.1080/13682199.2024.2398956.
[33] Branke, Jürgen, Kalyanmoy Deb, Henning Dierolf, and Matthias Osswald. 2004. Finding Knees in Multi-Objective Optimization. In *Parallel Problem Solving from Nature - PPSN VIII*, ed. Xin Yao, Edmund K. Burke, José A. Lozano, Jim Smith, Juan Julián Merelo-Guervós, John A. Bullinaria, Jonathan E. Rowe, Peter Tiňo, Ata Kabán, and Hans-Paul Schwefel, 722–731. Berlin, Heidelberg: Springer. doi:10.1007/978-3-540-30217-9_73.
[34] Augusto, Oscar Brito, Fouad Bennis, and Stephane Caro. 2012. A New Method for Decision Making in Multi-Objective Optimization Problems. *Pesquisa Operacional* 32 (August). Sociedade Brasileira de Pesquisa Operacional:331–369. doi:10.1590/S0101-74382012005000014.

**Figures**

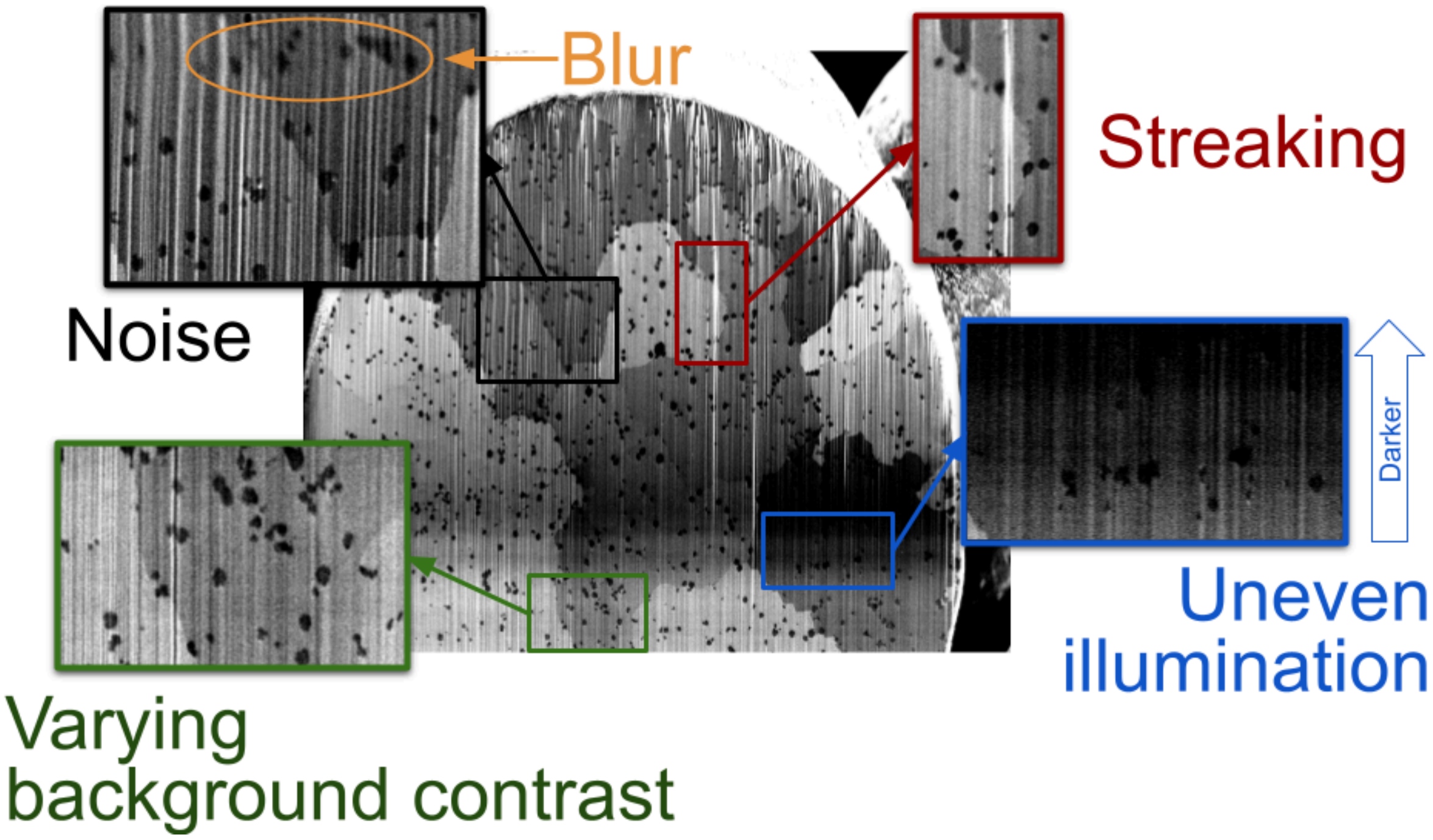

Figure 1. Sources of image degradation in electron micrographs of powder particles. *An electron micrograph of a GRCop-42 powder particle cross-section demonstrates challenges faced by processing algorithms: blur, noise, and uneven illumination arise from the electron microscopy process; varying background contrast by the composite nature of the alloy; and streaks formed when creating the cross-section.*

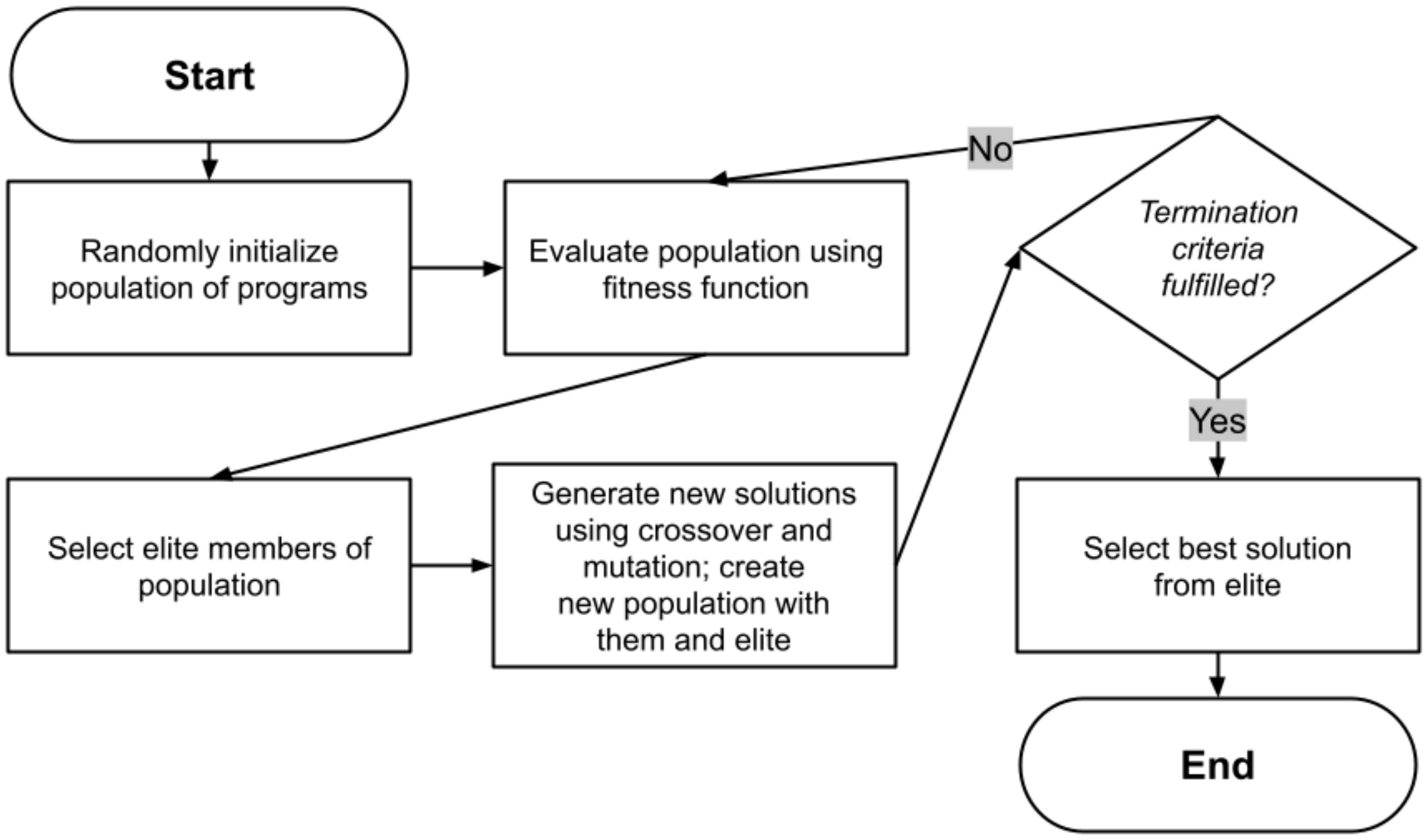

Figure 2. Optimization algorithm underlying genetic programming. *Genetic programming at its core is a powerful optimization framework because of its simplicity and flexibility. One only needs to implement five functions—program generation, fitness evaluation, selection, crossover, and mutation—and a small domain-specific language to implement the optimization algorithm above. This enables one to find satisfactory solutions to complex, nonlinear problems like performant image restoration algorithms.*

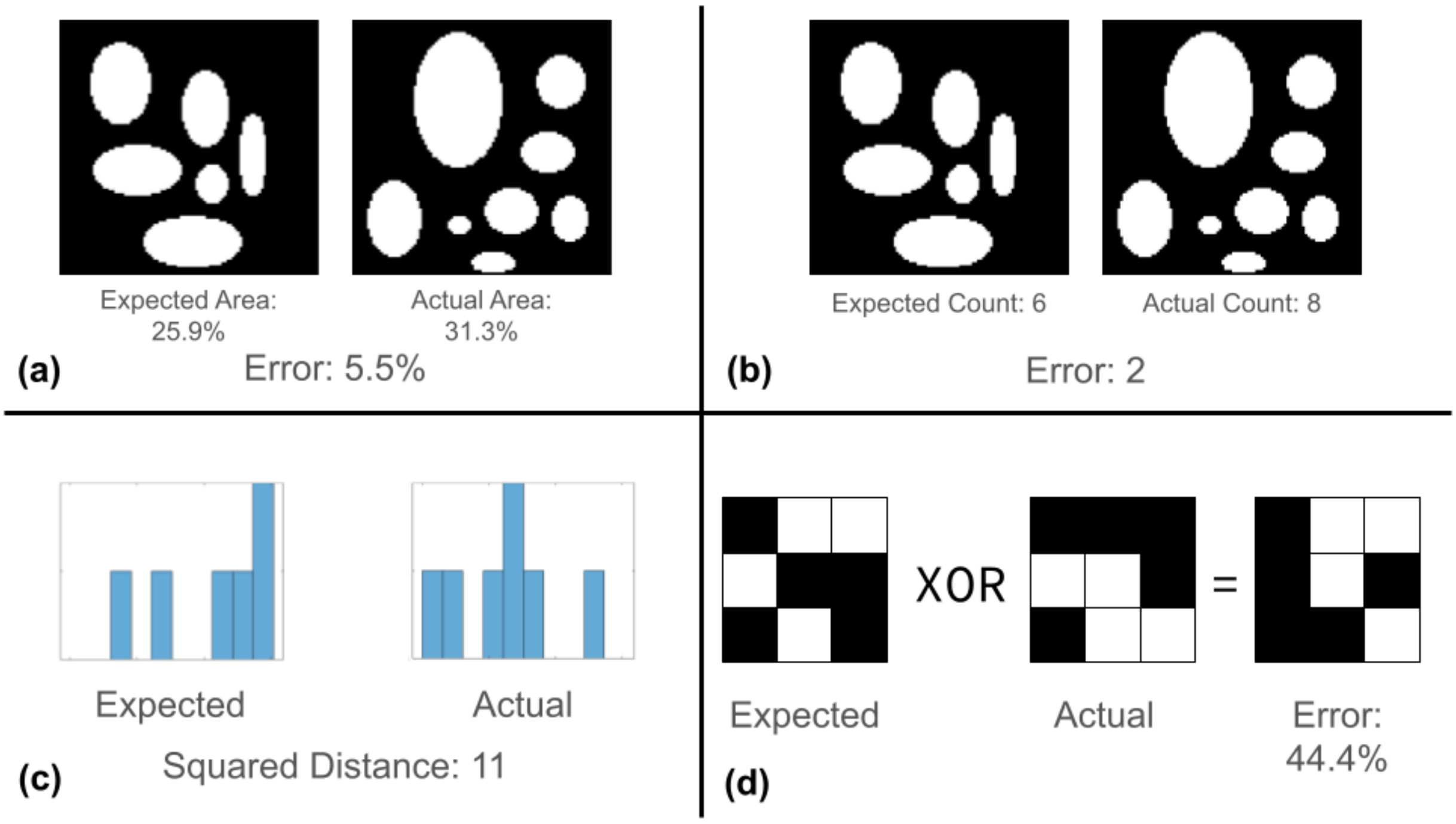

Figure 3. Objectives used to optimize image processing programs. *A visualization of the four surrogate metrics. Originally, when our environment's sole metric was XOR error, it overly rewarded algorithms that resulted in sparse, but ultimately inaccurate segmentations. Each of these metrics together address correctness from multiple levels of specificity in order to guide the environment towards performant algorithms.*

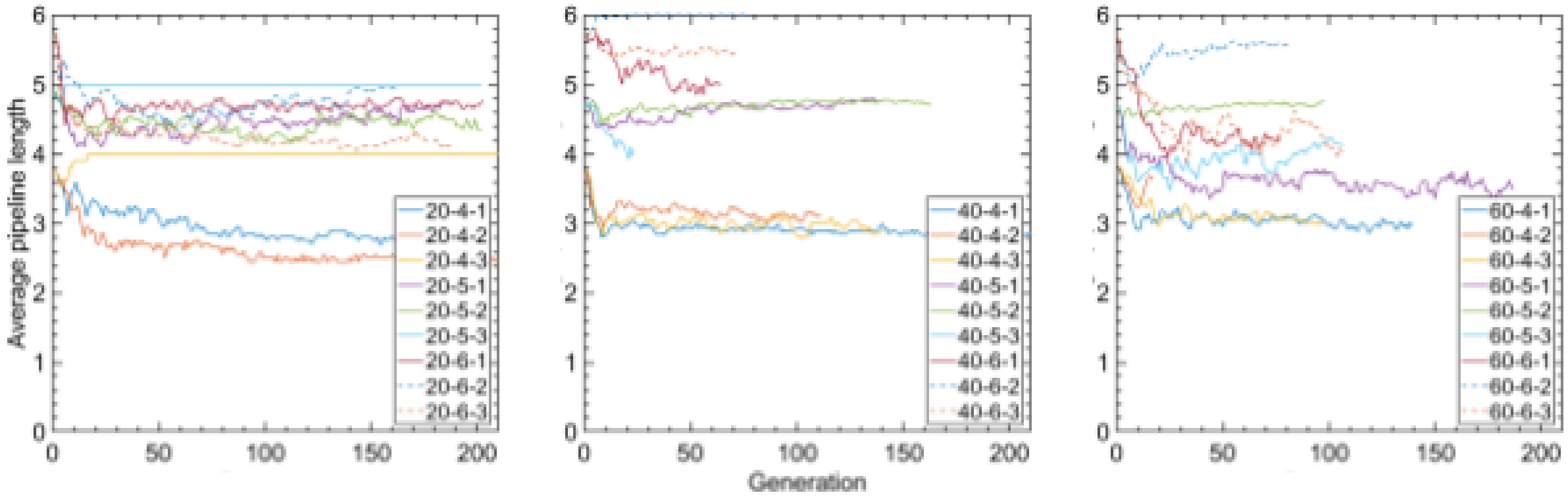

Figure 4. Average pipeline length versus generation vs trial. *Lines for each trial end at their final generation. Solutions generally tried using as few blocks as possible, although a few trials ultimately removed the identity block from circulation.*

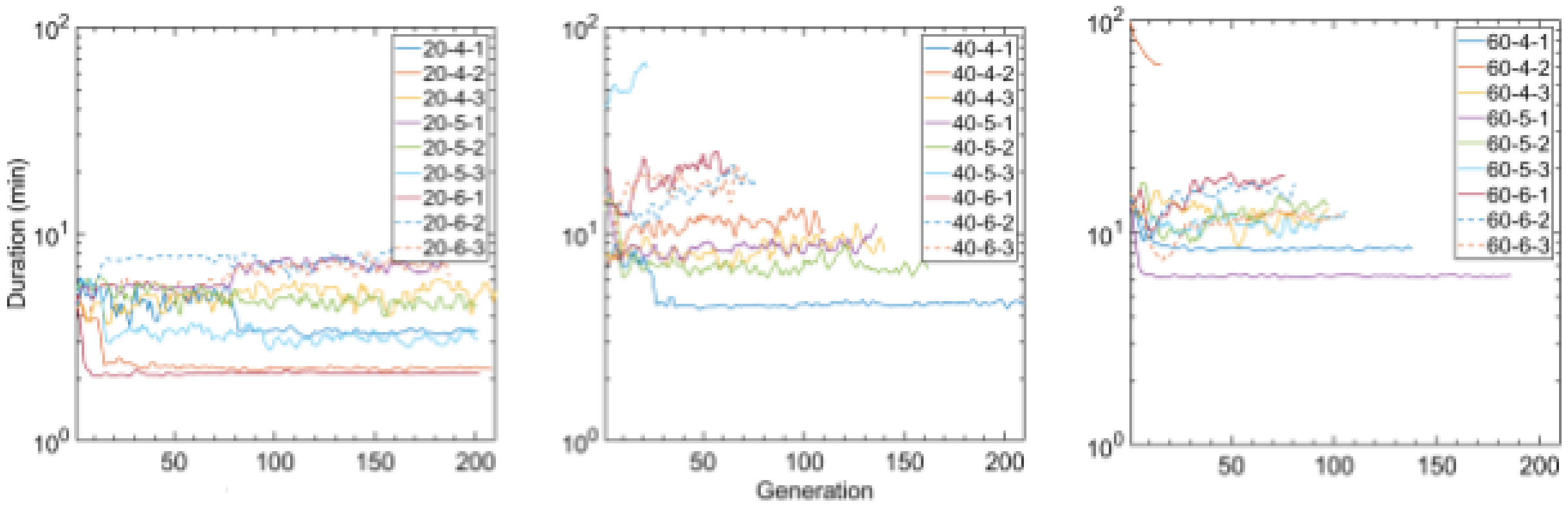

Figure 5. Generation duration versus generation versus trial. *Lines for each trial end at their final generation. A length-3 moving mean was applied to data for clarity. The general decrease and eventual plateau in generation duration shows that having time taken be a fitness metric was successful.*

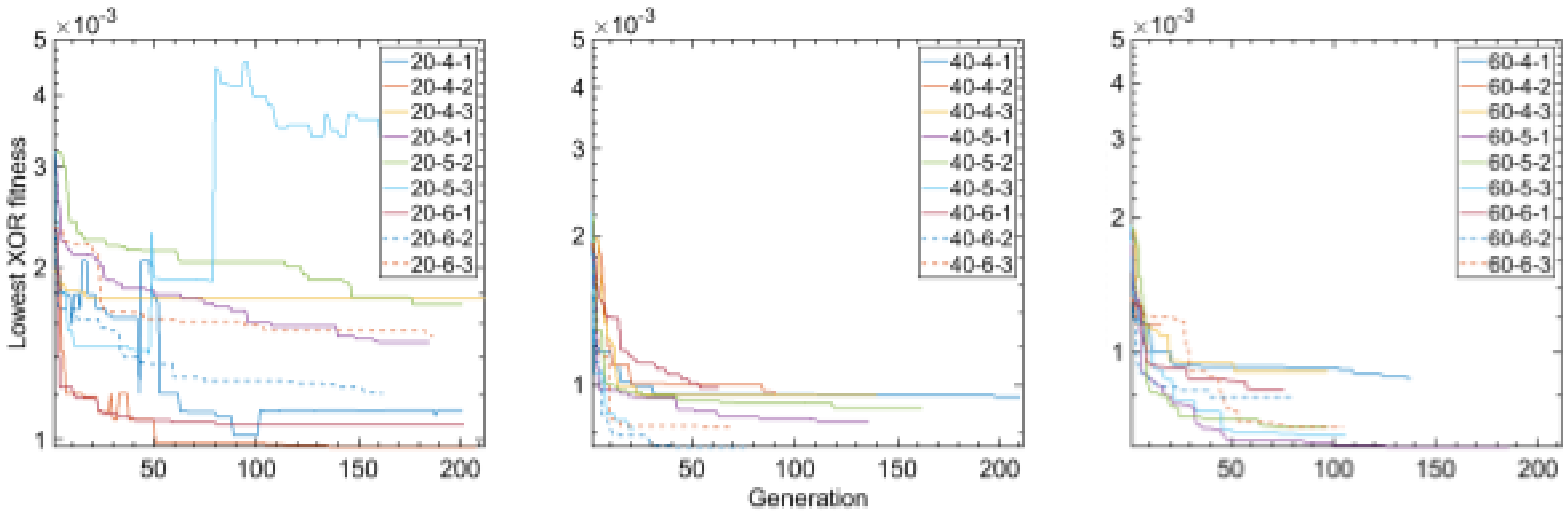

Figure 6. Lowest population XOR fitness vs generation. *Lines for each trial end at their final generation. Each line demonstrates the lower bound for performance shown by the population on the training set. The trials that were able to drive XOR error closest to zero are primarily those with populations of 60. Many trials plateaued in XOR fitness, focusing on other fitness metrics. Some trials, like 20-5-3 and 20-4-1 have sudden spikes in error.*

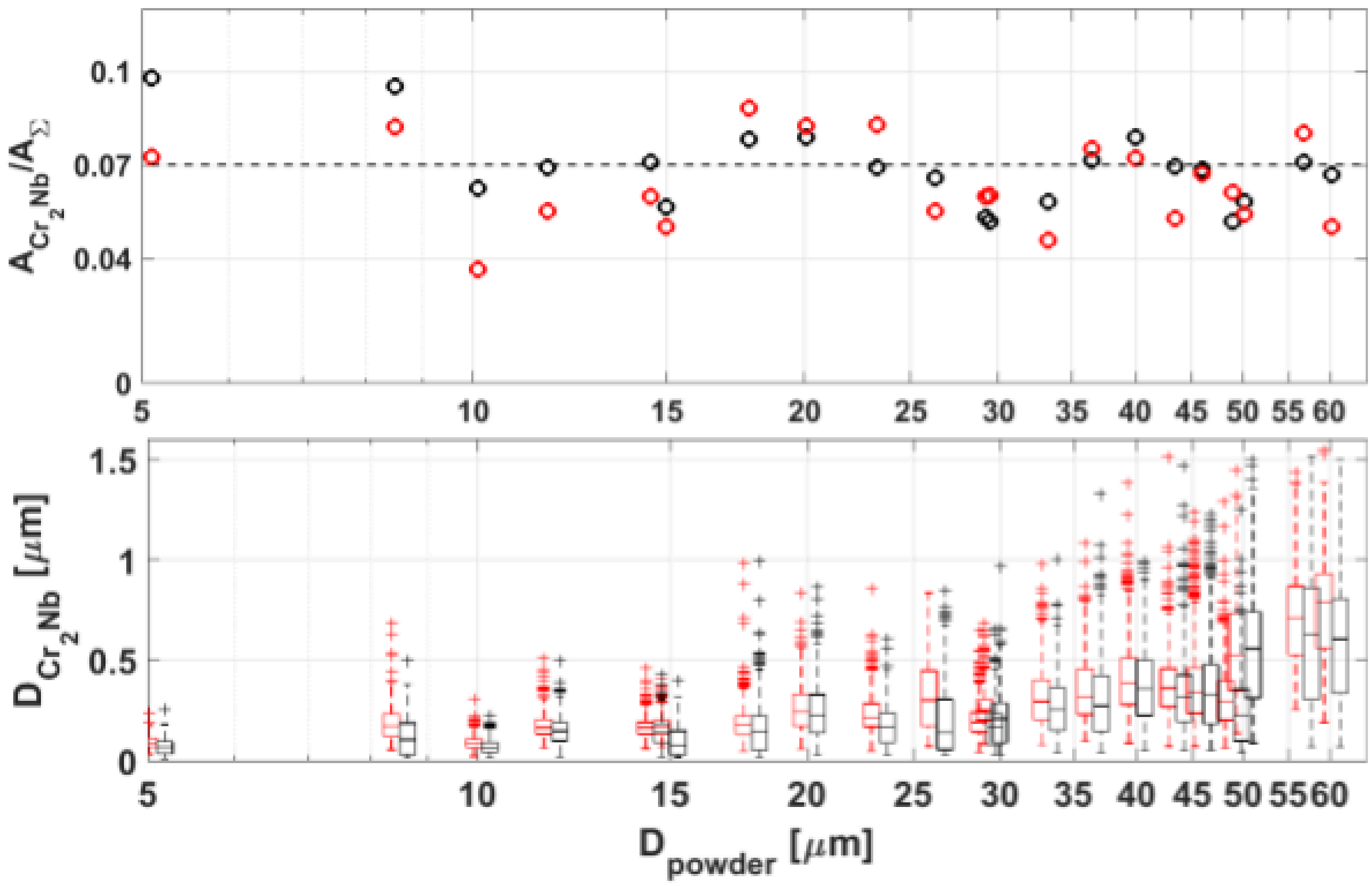

Figure 7. Comparison of computer-generated precipitate distributions to human baselines. *Human baseline is in black, while the computer is in red. For each micrograph, the top plot shows how much of the cross-sectional area is precipitate, while the bottom shows size distributions. For the bottom plot, for each precipitate diameter, the actual diameter is at the center of the computer-human box pairs.*

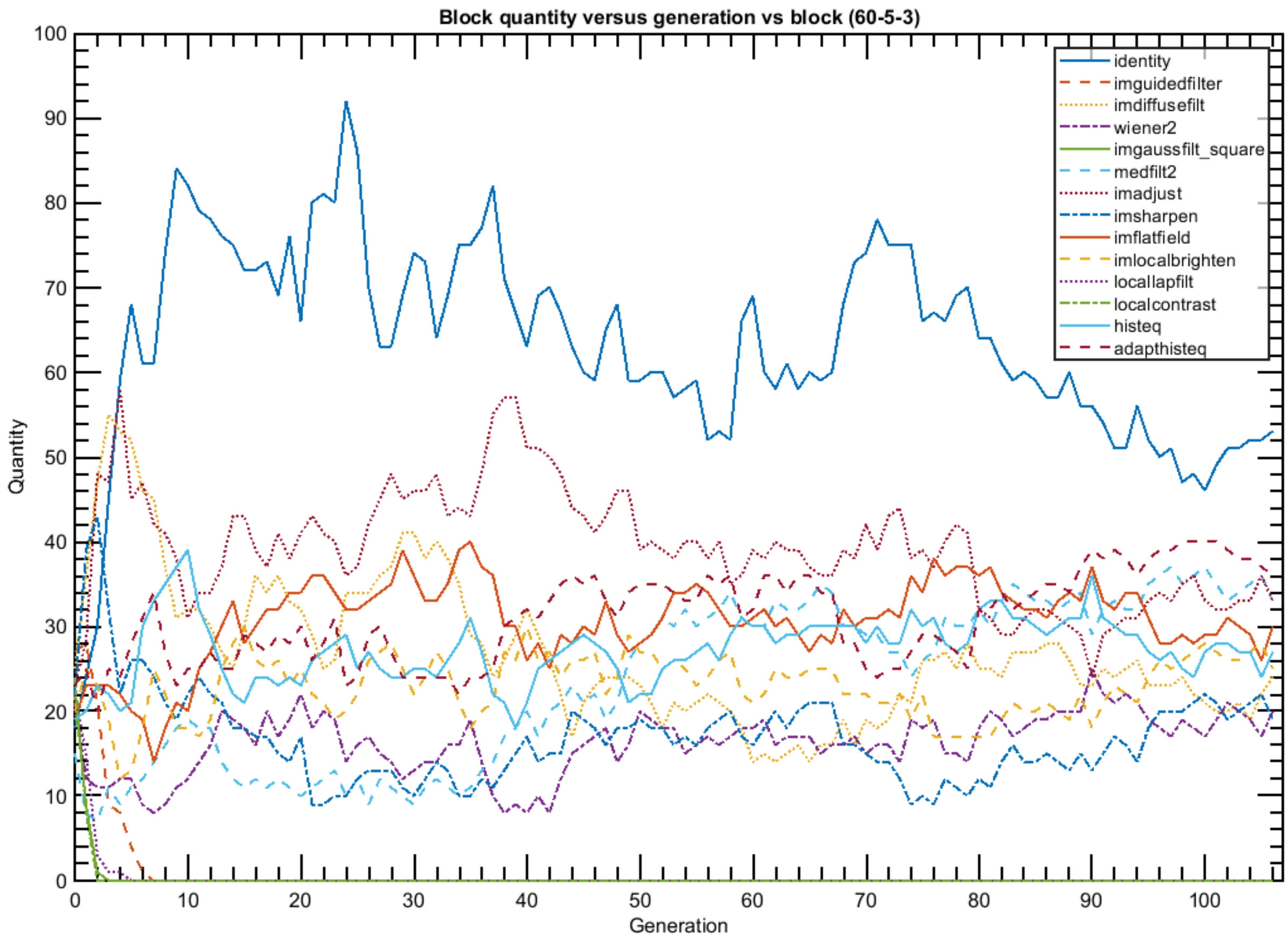

Figure 8. Popularity of blocks in a single trial across generations. *Graph of block frequency versus generation for a population for 60-5-3. Inefficient blocks "die out" quickly, while others vary in their frequency stochastically.*

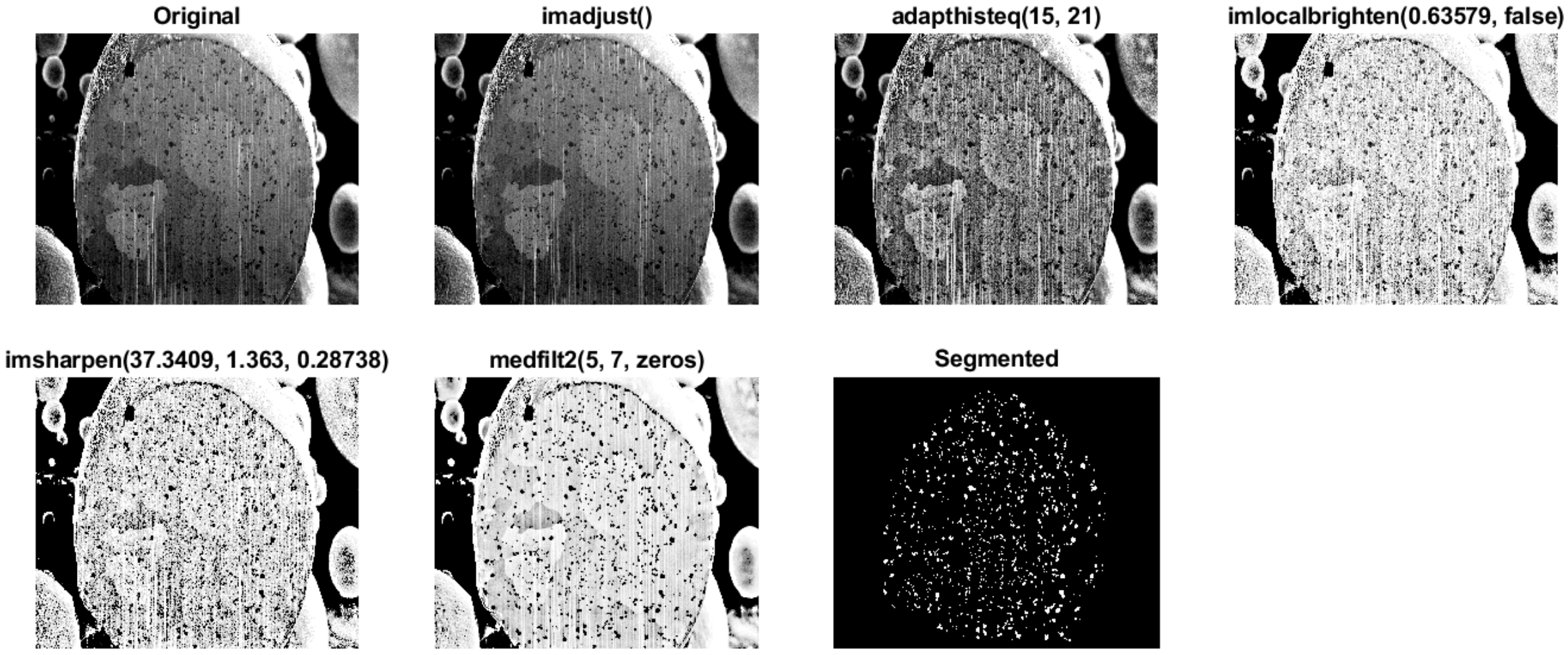

Figure 9. Demonstration of solution processing an electron micrograph. *Block-by-block pipeline filtering process of a micrograph with the most performant pipeline. An effect similar to flat-field correction is achieved without use of the* `imflatfield` *block. Blocks applied from left to right, top to bottom.*

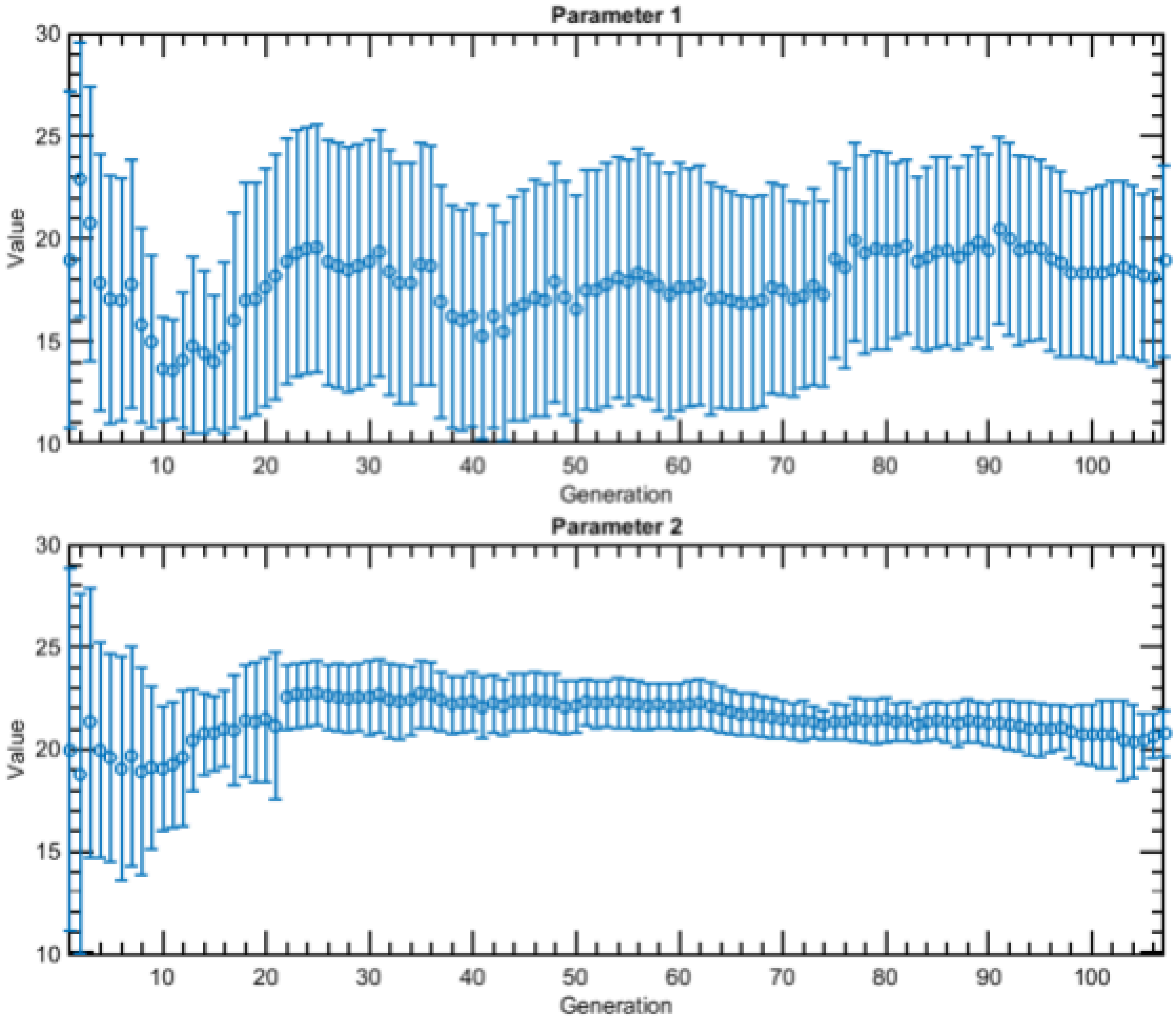

Figure 10.Parameters of `adapthisteq` over time. *Mean and standard deviation of parameter values as seen in the 60-5-3 trial. As time passed, the environment found a stable neighborhood height (Parameter 2) of about 21, but experimented much with neighborhood width (Parameter 1). Optimal neighborhood size for any one image is best found through experimentation; this implies that the environment may be trying to account for the wide variation in the training set.*

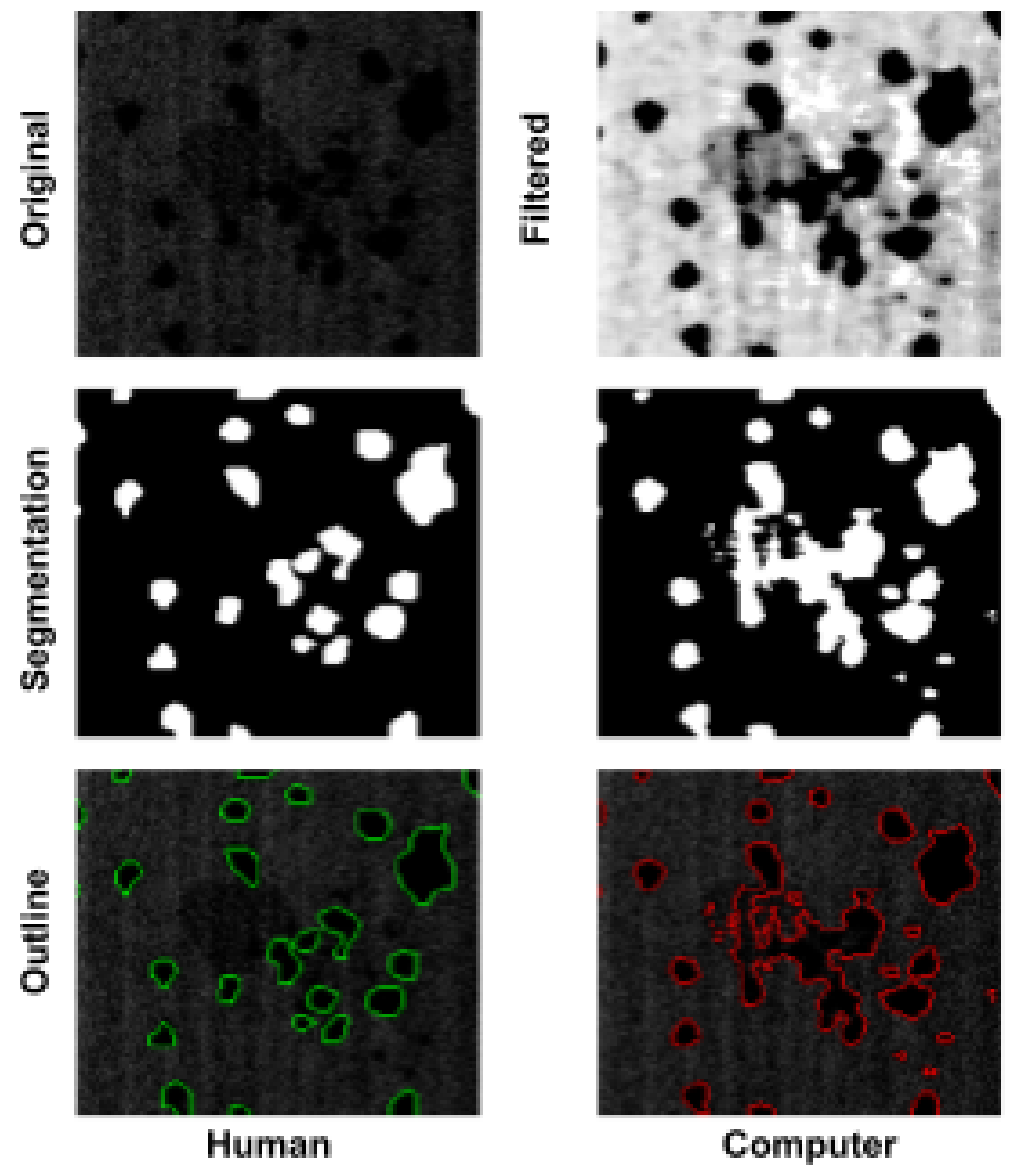

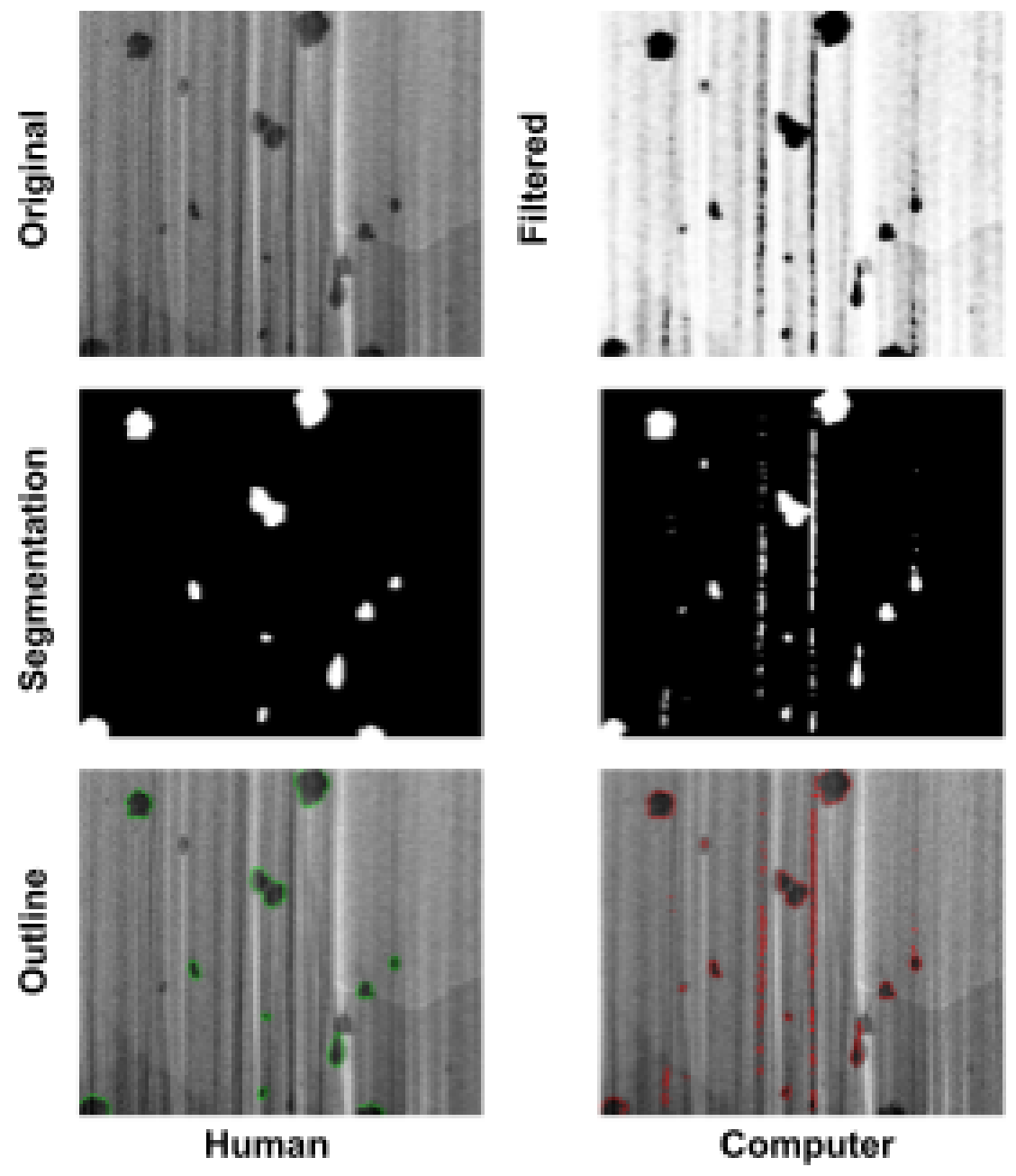

| | |
|---|---|
| Figure 11: Example of blobbing failure mode. *Our 60-5-3 solution, due to blurring caused by the median filter, connects the boundaries of multiple precipitates, causing them to be segmented by the computer all as one object.* | Figure 12: Example of streaking failure mode. *The algorithm has heavily amplified the stripe noise and made it contrast strongly with the background, effectively creating a "phantom" precipitate which is then segmented.* |

*From left to right, top to bottom: original micrograph, micrograph as filtered by our successful 60-5-3 pipeline, baseline segmentation of the original micrograph, segmentation of the filtered image by the pipeline, baseline segmentation overlaid on the original image, and same but for the pipeline segmentation.*

**Tables**

Table I. Minimum Average Evaluation XOR Error vs Maximum Program Length, Population Size.

| | 20 programs | | 40 programs | | 60 programs | |
|---|---|---|---|---|---|---|
| 4 blocks | Trial 1 | 2.2776% | Trial 1 | 2.1954% | Trial 1 | 2.103% |
| | Trial 2 | 2.1561% | Trial 2 | 2.078% | Trial 2 | 2.3797% |
| | Trial 3 | 4.5916% | Trial 3 | 2.0861% | Trial 3 | 2.1015% |
| | **Average** | 2.9996% | **Average** | 2.1198% | **Average** | 2.1947% |
| | **Std dev** | 1.3571% | **Std dev** | 0.0655% | **Std dev** | 0.1601% |
| 5 blocks | Trial 1 | 2.8189% | Trial 1 | 2.0823% | Trial 1 | 1.8002% |
| | Trial 2 | 2.9525% | Trial 2 | 2.4909% | Trial 2 | 1.8369% |
| | Trial 3 | 4.5651% | Trial 3 | 1.9917% | Trial 3 | 1.79% |
| | **Average** | 3.4455% | **Average** | 2.1883% | **Average** | 1.8090% |
| | **Std dev** | 0.9719% | **Std dev** | 0.2659% | **Std dev** | 0.0246% |
| 6 blocks | Trial 1 | 3.1043% | Trial 1 | 2.7298% | Trial 1 | 2.8675% |
| | Trial 2 | 2.568% | Trial 2 | 2.2336% | Trial 2 | 2.0524% |
| | Trial 3 | 2.912% | Trial 3 | 2.5366% | Trial 3 | 3.2461% |
| | **Average** | 2.8614% | **Average** | 2.5% | **Average** | 2.722% |
| | **Std dev** | 0.2717% | **Std dev** | 0.2501% | **Std dev** | 0.6100% |